# The Role of Robotics in Infectious Disease Crises

**Workshop Final Report**

October 2020

**A Workshop Organized by**

Gregory Hager, The Johns Hopkins University
Vijay Kumar, The University of Pennsylvania
Robin Murphy, Texas A&M University
Daniela Rus, Massachusetts Institute of Technology
Russell Taylor, The Johns Hopkins University

**With Support From**

Guru Madhavan, National Academy of Engineering
Ann Schwartz Drobnis, Computing Community Consortium

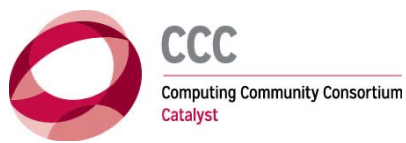

The material is based upon work supported by the National Science Foundation under Grant No. 1734706. Any opinions, findings, and conclusions or recommendations expressed in this material are those of the authors and do not necessarily reflect the views of the National Science Foundation or the National Academy of Engineering.

# The Role of Robotics in Infectious Disease Crises

## Overview

The recent coronavirus pandemic has highlighted the challenges faced by the healthcare, public safety, and economic systems when confronted with a surge in patients that need access to healthcare facilities and a population that must be quarantined or shelter in place. The most obvious and pressing challenge is taking care of acutely ill patients while reducing the risk of transmission within healthcare facilities , but this is just the tip of the iceberg if we consider what could be done to prepare in advance for future pandemics. Beyond the obvious need for strengthening medical knowledge and preparedness, there is a complementary need to anticipate and address the engineering challenges associated with infectious disease emergencies.

Robotic technologies are inherently programmable, and robotic systems have been adapted and deployed, to some extent, in the current crisis for such purposes as transport, logistics, and disinfection. Unlike many purely software applications, robotic systems inherently involve interactions with the physical world. Robots, just like other complex systems such as cars, airplanes, and manufacturing facilities, require longer times to adapt, deploy, and make reliable than many software applications. As technical capabilities advance and as the installed base of robotic systems increases in the future, they could play a much more significant role in future crises, especially if an appropriate strategy can be developed and implemented now that ensures adaptable and reliable robots are available or can be quickly replicated and distributed on demand.

Robots, just like other complex systems such as cars, airplanes, and manufacturing facilities, require longer times to adapt, deploy, and make reliable than many software applications. As technical capabilities advance and as the installed base of robotic systems increases in the future, they could play a much more significant role in future crises, especially if an appropriate strategy can be developed and implemented now that ensures adaptable and reliable robots are available or can be quickly replicated and distributed on demand.

In an effort to prepare for the next pandemic and perhaps aid in the current one, a virtual workshop entitled *Role of Robotics in Infectious Disease Crises* was co-hosted by the National Academy of Engineering (NAE) and the Computing Community Consortium (CCC). The workshop was held on July 9-10, 2020 and consisted of over forty participants including representatives from the engineering/robotics community, clinicians, critical care workers, public health and safety experts, and emergency responders.

This report is the outcome of the workshop and outlines a strategy for increasing national preparedness to use robotic systems and technology in future infectious disease emergencies. It identifies key challenges faced by healthcare personell and the general population and then identifies robotic/technological responses to these challenges. Then it identifies the key research/knowledge barriers that need to be addressed in developing effective, scalable solutions.

Finally, the report ends with the following recommendations on how to implement this strategy:

- **Conduct a full NAE consensus study** on the potential for robotic systems to assist in healthcare emergencies and to develop a National Strategy for increasing national preparedness to use robotic systems and technology in future emergencies.

- **Conduct scenario planning exercises** to "game out" future pandemic crises in order to identify areas where better preparation can facilitate more effective deployment of robotic systems to meet emergent needs.

- **Increase research addressing knowledge barriers affecting the capability and mobilization potential** of robotic systems to mitigate risk of communicable disease transmission.

# The Role of Robotics in Infectious Disease Crises

- **Initiate targeted research programs addressing environment-specific knowledge barriers** to the deployment of robotic systems in infectious disease settings**.**

- **Develop incentives for translational research to develop novel robotic systems for healthcare applications.**

- **Institute economic and policy incentives to accelerate adoption of adaptable robotic systems into everyday use by public agencies, particularly healthcare and the medical supply chain industry.**

- **Establish a consortium focused on robotic systems in infectious-diseases** to facilitate connections between roboticists, government, industry, and citizen stakeholders and to provide a clearing house/repository for validated system designs and solutions that can be shared within the community.

# The Role of Robotics in Infectious Disease Crises

## 1. Identify key challenges faced by health care responders and the general population

Infectious disease at a national or global scale impacts people's lives in innumerable ways, some of which are immediately obvious -- for example, they require evaluation and management for the disease itself -- and some of which are indirect -- for example, their normal work-related activities are curtailed to limit person-to-person spread. As such, it is useful to view the impact of disease as emanating from two epicenters -- one oriented around acute evaluation and management for the disease itself, and one from the public health implications of the disease. The remainder of this section is organized around these two themes, calling out the many direct and indirect impacts that were surfaced during the workshop.

### Acute Care

A pandemic imposes an immediate and highly disruptive burden on the health care system. Like many industries, healthcare has been steadily driven toward efficiency through optimization and specialization of operation. As such, hospitals are increasingly designed to operate with a certain patient volume and mixture. Operating rooms are filled by patients undergoing elective and non-elective procedures, , rehab facilities are filled with convalescent patients, and intensive care unit (ICU) capacity is designed to service the expected needs. Staffing, food, medicine, personal protective equipment, and other supplies are stocked to meet this demand, and local, national, and global supply chains are in turn optimized to meet nominal demand.

A pandemic such as COVID-19 disrupts normal operations by placing an unsustainable burden on a small set of care pathways within healthcare, while at the same time attenuating or eliminating many of the other operations needed to support the ongoing healthcare needs of the uninfected population.

#### Disease-related care

Many of the challenges that arise in disease-related care are immediate consequences of the need to rapidly scale a normally limited set of operations to a much larger population. To achieve this scaling, a healthcare operation has to quickly expand 1) the available staff to support patient diagnosis and treatment; 2) the facilities within which patients are treated (i.e., beds, with specific capabilities); and 3) the supplies needed to treat them.

The evaluation and management of an emerging infectious disease such as COVID-19 places enormous new demands on staff. All treatment must take place with protection in place for both caregiver and patient. The donning and doffing of personal protective equipment (PPE) slows turnover between patients ; each entry and exit requires donning and doffing PPE, respectively. a patient room to check their condition a complex infection control process. This is happening as patient counts are growing, each care-giver is responsible for an increasing patient load, and the monitoring and care needs for patients are especially acute [1] In aggregate, the number of actions performed by every care-giver grows exponentially (and therefore unsustainably) with each new patient.

Surrounding this are increased needs for facilities support, particularly infection control. Furthermore, patient length of stay is increased, thereby increasing the overall population which in turn increases the need to move supplies -- food, medication, bedding and so forth -- throughout the hospital. Not only does this create immediate logistical challenges, but it also challenges the overall inventory management of the hospital itself.

# The Role of Robotics in Infectious Disease Crises

Finally, it is important to note that as patients recover, there are down-stream impacts such as the need to support convalescent treatment and rehabilitation both within and outside of the patient settings. These important services are in danger of being eclipsed by the front-line acute care needs.

**Broader impacts**

During a pandemic, attention naturally turns to the immediate challenges of disease diagnosis and treatment. However, the broader range of healthcare needs does not stop -- people still have strokes or heart attacks, suffer accidents, or have chronic conditions that require management. During the most recent pandemic, two trends have emerged. The first is a remarkable drop in the number of patients presenting for typical healthcare needs; the second is the emergence of telemedicine as a surrogate for traditional office visits. It is worth noting the former has a number of immediate financial implications for hospitals [2] as well as broader implications for public health [3].

Telemedicine is an area that provides much needed elasticity in healthcare capacity. However, current telemedicine options are still relatively rudimentary, particularly where direct physical measurements, such as vitial signs, are required and/or remote physical manipulation is needed [4]. While there are a number of promising technologies already available or in the offing for remote biometric measurement, both regulatory, logistical, and systemic impediments have limited the development and integration of these technologies into routine patient care.

Finally, it is worth noting that almost every aspect of hospital operations are affected or disrupted during a pandemic. For example, the number and type of lab assays will likely change during an ID event. As with front-line care, the equipment and staff needed to service this change in demand is not elastic and thus places unique demands on healthcare infrastructure.

## Public Health, Safety and Welfare

Beyond the acute care setting, an pandemics produce widespread disruptions resulting from policies implemented to control the spread of disease. In an extreme case, such as with COVID-19, these impacts are pervasive throughout society. Here we focus on articulating some of the broader themes that are related to management of infectious disease spread specifically as well as societal implications.

**Control and Spread**

The Center for Disease Control (CDC) has a number of recommendations to control of infectious diseases -- reducing transmission by early detection, isolation, and contact tracing to identify and non-pharmaceutical interventions such as masking, hand hygiene, disinfection, and physical distancing, restrictions on movement and size of gatherings and other activities to  mitigate transmission of disease [5].

With respect to reducing transmission, the need for improved disinfection methods goes well beyond healthcare institutions.  Transport centers, restaurants, and entertainment areas all will need to move to a new heightened level of cleaning. Similarly, there is a need for improved monitoring for surveillance for suspected disease -- elevated temperature, coughing, sneezing, and so forth. Ideally, many of these mechanisms would eventually "trickle down" into individual homes. Early identification of disease in the broader population also requires the deployment of large-scale testing and/or contact tracking to quench disease spread. This presents both logistical challenges as well as placing necessary health care workers at increased exposure risk. Finally, production and distribution of elements for controlling disease spread -- from PPE to disinfection agents to vaccines, requires a substantial retooling of supply chains. In particular, it is worth noting that for a global pandemic such as COVID-19, it will be necessary to produce, distribute, and administer billions of doses of vaccine. This will require massive adjustments to production facilities as well as new ways to move materials to

# The Role of Robotics in Infectious Disease Crises

the front line and support deployment.

**Broader impacts**

As we have seen with COVID-19, the potential impacts of a pandemic, if not controlled early, are deep and pervasive. Many of the challenges faced in the healthcare setting -- supply chain disruptions, the change in demand, and the need to support remote work are reflected in much broader economic settings. However, the impacts extend beyond this.

One of the unique challenges we have seen in the current pandemic is the dual stress of changes to supply chains -- from food production to supplies to waste removal (e.g., [6]) -- combined with the disease impacting these supply chains ability to deliver. In some cases, as with telehealth, this has driven industries to rapidly investigate remote work options, or accelerate other forms of automation.

However, industries that demand physical presence -- from packing plants to distribution centers to garbage pickup -- are currently extremely limited in what they can accomplish remotely/virtually and are largely relying on increased surveillance, masking, distancing, and increased frequency of disinfection as mitigation strategies. Beyond this, there are clearly parts of society who have borne disproportionate burden of both disease, economic impact, social isolation, and other negative impacts.

## 2. Identify robotic/technological responses to these challenges

There are many ways in which AI and Robotic technologies augment the response to a pandemic [7]. There is also great potential in using technologies in leveraging machine learning, data science and automation in the research and development leading up to the development of the vaccine, and the manufacturing, and the supply chain for distribution. But more immediately, robotic technologies can provide hospital staff with tools to safely treat patients, remotely and effectively. Robotic technologies also have the ability to support patients by providing an interface with the outside world that can include family members and friends on one hand, and health care professionals on the other. Robots have historically played an important role in tasks that have been described as dull, dangerous and dirty. There are many examples of tasks like these in testing, monitoring, and treatment and care of patients that are candidates for robotics. They can also support the patients psychologically by facilitating telepresence with family and friends. We provide a few specific examples next.

First, **telepresence** has been touted as an important technology for over two decades but primarily in the context of underserved rural communities which may not have access to specialists and doctors. It is now recognized that the infrastructure required in some cases may be significant and the logistics challenges may be hard to overcome. But the challenges associated with urban communities and with healthcare centers are simpler. Telepresence robots that have very little autonomy (e.g., iPads on remotely controllable mobile bases for video conferencing) allow specialists to visit patients remotely maximizing efficiency and safety.

Robotic systems can help with direct patient support and conserve resources in short supplies such as masks, gloves, and gowns. For example, teleoperated robotic systems can facilitate operation of equipment such as ventilators or infusion pumps without requiring ICU personnel to don PPE to enter the room for routine adjustments.

# The Role of Robotics in Infectious Disease Crises

For example, robots carrying touchless sensors for vital signs can collect the vital measurements from the patients. Robotics can increase the capacity of the system if it is able to perform some of the patient management functions (e.g., changing IV bags, managing a ventilator) without requiring a caregiver to move into and out of the room. These may require **semi-autonomous mobile manipulation**, potentially with shared control, and allow Hospital Clinical Personnel (HCP) to perform a multitude of tasks without entering the patient room. They can also perform many supply chain functions within healthcare facilities.  For example we can ensure that special equipment and medicine gets delivered just on time to patients and procedure rooms.

There are many tasks that are currently performed by HCP that can be performed by robots. This includes cleaning and disinfection tasks.Indeed **cleaning and disinfecting robots** could be as essential to hospitals as MRI or heart-lung machines. It is important to recognize the challenge in developing robots like these since they will be required to operate in environments that are unstructured and will require a significant level of autonomy. Autonomous robots may also make it possible to avoid use of tedious, time-consuming techniques (e.g., manual disinfection in complex settings) in hospitals, and they can also facilitate automated record-keeping and audit trails for these tasks.

There are many opportunities for using robots in **public health support functions** that are not directly related to care in hospitals. For example, intake into the hospital almost always requires some level of triage and/or testing. This is again skilled labor intensive, and, in the case of communicable diseases, creates the potential for transmission. Robotic systems are ideally suited to perform many of these initial triage functions and can be used outside of facilities, including in rural locations where there are few health workers. Robots can take temperature, blood pressure, capture sounds such as coughing and can even perform basic imaging functions (e.g. ultrasound) and perform sample collection (e.g., nose swabbing, phlebotomy). This may improve the efficiency of evaluation of patients while mitigating risk of potential exposure to HCP and conserving use of PPE.  Similarly, continuing  advances in sensor technology can assist in detection of pathogens.  Such technology might be incorporated into robotic systems for disinfection or surveillance or might also be incorporated into inexpensive materials.  For example, one might imagine a face mask that lights up when a pathogen is detected.

During the COVID-19 pandemic we saw a significant shift of health care from hospitals and clinics to homes as to both deliver care where it was most appropriate as well as decrease demand on acute care and post-acute care facilities. The need to monitor patients after discharge or doctors' visits is also significant. Patients are more likely to be isolated due to "shelter in place" orders and because they are disconnected from their support network, and even because of loss of employment. The increased need for home care, rehabilitation and mental health care suggests a tremendous opportunity for **home care robots,** which have long been considered viable by the research community, but with the potential now to provide companionship to alleviate the effects of isolation and distancing.

## 3. Identify key research/knowledge barriers that need to be addressed in developing effective, scalable solution

As highly programmable devices, robotic systems can be adapted to function effectively in widely varying and often unstructured environments. Continuing advances in rapid prototyping, sensing, machine vision, and artificial intelligence will further increase this adaptability.

However, significant barriers remain to the development of robust and physically dexterous robotic systems that can function effectively in infectious disease environments with efficiencies approaching

# The Role of Robotics in Infectious Disease Crises

that of humans. For the purposes of this discussion, we will divide these barriers into three broad categories:

- **Capability barriers**, i.e., research and knowledge barriers common to the development of intelligent robotic systems capable of performing useful tasks in our broader society.

- **Mobilization barriers**, i.e., research and knowledge barriers associated with the potential need for rapid development or redeployment of robotic systems customized to meet the immediate needs of an emerging crisis.

- **Infectious disease environment-specific barriers**, i.e., research and knowledge barriers associated with the challenges of placing robotic systems in highly infectious or contaminated environments.

We will discuss many of the associated infrastructure issues in Section 4.

## Capability Barriers

**Autonomy and Intelligent Systems:** Many potential uses for robotic systems in responding to pandemic crises require that the systems operate autonomously or semi-autonomously in unstructured environments. Broadly, there are two fundamental questions: 1) How can the robot be told unambiguously what it is supposed to accomplish. And 2) How can the robot reliably and safely carry out its assigned task. For simple behaviors, there are existing solutions such as teleoperation. But the challenges increase markedly as the level of autonomy increases. Significant advances are needed in all of the underlying components associated with intelligent systems. These include:

- Combining information from multiple sensors and sources to create and maintain an effective model of the task environment.
- Methods for specifying tasks that are "understood" both by the robot and by humans.
- Methods for the robot to plan and control its actions to accomplish the task, based on an evolving model of its environment.
- Methods for the robot to detect unexpected events and to make appropriate responses in a safe manner.

These capabilities are especially important if the robot is to work in an environment that was designed for human users rather than specifically for the robot.

For example, consider a mobile robot designed to perform simple tasks in an ICU, such as monitoring and adjusting ventilators and infusion pumps, changing IV bags, or the like. The robot has computer vision capabilities, a mobile base, and two manipulator arms. Although the robot may often be in the room alone, clinical staff also perform other tasks in the room and equipment and the patient bed may change or be moved from time to time. Suppose that the immediate task is to check on an infusion pump, adjust the flow, and change a bag if needed. The robot could use computer vision to identify and locate everything in the room and traverse the room to the pump. Like most "legacy" equipment, the pump is unlikely to have a networked interface, but the robot can use computer vision to read the information displayed on the pump controller and compare the readings to some pre-specified limits.

Based on what it sees, the robot may use its manipulators to adjust the settings or communicate via wireless network to a human expert if the situation is unusual. The robot can also use its vision system to assess the contents of the associated IV bags and uses its manipulators to change the bags and reconnect them. The robot also needs the ability to detect unexpected events (such as missing equipment) and either to report them or take corrective actions itself. Doing these things in a robust and general manner would be extraordinarily challenging or infeasible for current-day systems.

# The Role of Robotics in Infectious Disease Crises

For another example, consider a robot for cleaning and disinfection of surfaces and equipment. Although there are existing mobile systems for cleaning floors and for using UV light for disinfecting spaces, both in hospitals and in public areas like grocery stores, the problems associated with cleaning which occurs in conjunction with disinfection, are much harder. The robot needs to recognize objects, determine what can be cleaned and what should not be disturbed, physically interact with and possibly move the things to be cleaned and disinfected, and make real time decisions about how to adapt its general capabilities to the individual circumstances of the task at hand.

**Human-robot interfaces and interaction methods:** Many of the robotic applications that will be most useful in a pandemic must necessarily involve interaction with people, both the individuals charged with supervising what the robots are supposed to do and others who may simply be in the environment or whom the robot is trying to help. These people will **not** be expert programmers or Ph.D. computer scientists. Many will have little or no real technical training, beyond perhaps some familiarity with smart phones or social media apps. But they must communicate with the systems and cooperate with them without being intimidated. It is very important that there be mutual trust and some shared understanding between the robotic systems and the human users. The humans must understand what the robot will do and have some confidence that it is doing it. The robot must "understand" what the human expects it to do and have some reasonable expectation of how the human will respond. It must be able to react appropriately if the human does something unexpected.

Not nearly enough is known about either the "social" or the "technical" aspects of human-robot communication and cooperation. Significant advances are needed in these areas, including:

- Speech and natural language processing
- Haptics and physical interaction between humans and robotic systems
- Management of task-oriented interactions between robots and humans
- Modeling of human expectations in human-robot interactions
- Methods for "inexpert" users to specify desired tasks and constraints to robotic systems
- Methods for giving "non- expert" users necessary information about the robot's task performance
- Methods for rapid training and development of human-robot teamwork involving novice users.

**Information science and communications:** We live in an increasingly networked and information-intensive world. The ability of robotic systems to communicate with each other, with large data bases, and with information processing systems can be critical both in crises and in more normal societal situations. A few areas of special synergy with robotics include:

- Methods for gathering, organizing, and summarizing enormous quantities of sensory data in ways that can be used to facilitate situational awareness and machine learning.
- Methods to exploit the sensing and mobility capability of robotic systems to perform environmental monitoring and other "surveillance" tasks.
- Methods to record and learn from task execution in the course of routine activity.
- Methods to preserve privacy while doing the above.

**Hardware and physical capabilities:** In addition to the computational capabilities described above, further advances in the underlying manipulation and sensory components of robotic systems are needed. Examples include:

# The Role of Robotics in Infectious Disease Crises

- Sensor technology, including optical, touch, force, haptic, and other sensors that can be readily incorporated into robotic systems.
- "Soft" robotic actuators and mechanisms
- Low-power and energy efficient actuators and sensors
- Compact, high dexterity mechanisms

## Mobilization Barriers

The deployment of robotic systems to respond to a pandemic will necessarily present many challenges, even if the needed technical capabilities are already available in some existing systems. Just as in the current COVID-19 pandemic, the response to the next crisis will doubtless include some combination of repurposing/reprogramming existing systems and the rapid development and deployment of new systems to meet particular needs. There are both knowledge and infrastructure barriers that need to be addressed in order to enable a fast and robust response. Infrastructure needs will be discussed in Section 4. Here, we will focus on a few of the key knowledge barriers that seem particularly important in enabling a rapid response to a crisis. As with the "capability" barriers discussed above, these barriers are not specific to infectious disease crises, and advances in overcoming them will have immense benefits much more broadly for our society.

**System engineering:** Robots are systems in which component subsystems must interact in often complex ways. Modifying an existing system to change its programming or to add a component in order to meet a new need can present challenges in predicting how the changes may affect the overall system performance, and the difficulties increase as systems become more and more complex. These challenges are exacerbated in cases where an entirely new system is being developed and as "artificial intelligence" methods make some components more and more opaque. Significant advances are needed in all aspects of system engineering. Examples include:

- System and interface design specification and certification methods
- Design verification and test methods (see also below)
- Safety
- Fault detection, tolerance and recovery methods

Within this context, improved methods for **risk assessment, projection, mitigation** for complex systems deployed in broad societal contexts are especially important. Better formal methods for quantifying risks and effects of malfunctions in highly complex systems involving human-machine interaction in real-world scenarios would significantly enhance trust by stakeholders and facilitate the deployment of newly developed or adapted systems in crisis situations.

**Testing and Verification:** Testing and verification is vitally important for **safe and effective** deployment of complex systems that must interact with humans and perform useful work in relatively unstructured environments. It is crucial that testing include significant evaluation with non-expert users similar to those who will encounter the system in actual practice. Testing scenarios must cover both "normal" situations that the system is likely to encounter in practice, as well as "corner" cases that are unlikely but may occur.

As mentioned above, the introduction of "artificial intelligence" and "machine learning" methods into robotic systems have made these issues especially challenging. Training data must be well characterized and free of unconscious biases. The system must be able to recognize and respond

# The Role of Robotics in Infectious Disease Crises

appropriately to unexpected situations outside the span of the training data. Better formal methods for validating these systems are badly needed, as is the development of better testing procedures for evaluating their performance.

**Agile and low-cost manufacturing:** Deploying or modifying large numbers of robotic systems in time to make a difference in an emerging crisis will necessarily the ability to manufacture components and entire systems quickly, without excessive time spent designing and building specialized fixtures or assembly tools. Similarly, replication costs may severely limit the number of systems that can affordably be deployed, even if they meet a real need. The response to the current COVID-19 pandemic has shown the value of current-generation agile manufacturing technology, especially 3D printing. But further research advances to improve both the capabilities and reduce costs of these systems will significantly enhance our national readiness to respond to a future pandemic.

## Environment-Specific Barriers

The use of robotic systems presents some obvious challenges beyond those encountered in other sectors of our society. Although medical device manufacturers have extensive experience in developing devices for use in healthcare settings that are compatible with the variety of EPA-registered disinfectants for the specific use or pathogen, further research in several areas could significantly facilitate rapid development and deployment of systems in a crisis. For the purposes of this discussion, we consider that these robots are non-critical equipment per the Spaulding Classification Scheme and thus require low-level disinfection. For example:

- **Development of advanced materials that are inherently self-disinfecting** could greatly simplify design and deployment in a crisis, especially if these materials are inexpensive and compatible with 3D printing and other agile manufacturing processes. The development of materials that facilitate cleaning prior to disinfection also would be advantageous.

- **Development of novel sensors and sensing methods to detect infectious agents** that can be combined with robotic systems for contamination surveillance, cleaning, and disinfection.

- **Development of robotic systems that can assist clinical staff** in tasks such as patient handling, sample collection, and PPE donning/doffing. Although most of the capabilities associated with such systems are more generally useful, the unusually intimate robot-patient or robot-staff contact associated with such applications can present special problems.

- **Development of knowledge bases and communication channels** to promote mutual understanding between clinicians and other professionals dealing with communicable diseases and engineers who can develop robust, deployable systems to meet emerging needs. One of the main challenges in the current COVID-19 pandemic has been the time required to bring robotics researchers up to speed on the needs and constraints associated with working in infectious environments, together with a need to give user communities sufficient background in robots and technology in order for them to help develop and assess possible solutions.

## 4. Identify workforce training, regulatory, and infrastructure needs that should be addressed in order to enable rapid deployment of these systems

In responding to a pandemic crisis, simple rapid deployment of robotic systems may be counterproductive. The goal should be rapid, *responsible* deployment, where the robotic innovations do more good than harm, including secondary consequences to a broad set of stakeholders. Responsible robotics innovation is very different from the "something is better than nothing"

# The Role of Robotics in Infectious Disease Crises

approach that has been shown to be counterproductive for disasters. The introduction of robots have made some disasters worse. Likewise, some robots deployed for search and rescue have violated regulations and expectations of privacy. A recent report analyzing 203 instances of use of robots in the current COVID pandemic between March 2020 and July 2020 has already produced at least 45 expressions of ethical concerns over various applications [8].

There are at least three challenges in responsible robotics innovation for disasters, many of which have been identified as challenges for computing systems for disasters [9, 10]. One challenge is that disasters have a short opportunity for impact, meaning that reliable robots have to be immediately matched with stakeholder needs, i.e. *demand analysis*. A maxim in disaster response is that every incident presents a new, unforeseen situation. However, stakeholders (e.g., public agencies) often do not know what is possible and thus cannot articulate their emerging needs. Roboticists often do not have domain knowledge about applications so innovations may not be a good match. The challenge imposed by short time of opportunity is amplified by the years it takes to produce a reliable robot capable of a heavy duty cycle; errors in the demand analysis may lead to a robot that is not adopted.

A second challenge is the spectrum of domain and stakeholder complexity, especially for pandemics and outbreaks, which implies that multiple segments of the workforce need to be trained, that multiple regulations may be impacted, and information technology infrastructures will have to be expanded. The broad use of robots for COVID-19 illustrates that workforce, regulatory environment, and infrastructure issues go far beyond hospitals. Indeed, hospitals were not the major point of care in the 2014-2016 Ebola and the 2016 Zika outbreaks. The use cases for COVID-19 response includes public safety, public health, businesses, and individuals, with Public Safety having more reported instances of robot use than clinical care [15]. These application categories have different economic and regulatory eco-systems, including health insurers and business continuity insurers and competing regulations, standards, and at the federal, state, and local levels [8]. Fortunately, most of the regulatory agencies have procedures for waiving or expediting relaxation of regulations. The most stringent regulations, which are for medical devices, are possibly the most visible, but there are also relevant regulations for other application areas, as well.

A third challenge for responsible innovation for disasters is the trustworthiness of the robots in the sense of trusting the reliability of the robots or their suitability for the application. Stakeholders are risk averse and typically reject novel robotics innovations as untrustworthy for the intended application.

Unfortunately, roboticists do not have formal methods for projecting i) the risks of malfunction or unexpected behavior, ii) the impacts on the stakeholders work processes who are already stressed or working at capacity (e.g., extra work or increased fatigue and then need for time consuming training [11, 12], and iii) secondary consequences (e.g., environmental impacts of overspraying disinfectants to sanitize public spaces). While NASA has created two scales for technical maturity [13] -- technical readiness levels (TRL) for the reliability of the device as a unit, and technical readiness assessment (TRA) for the suitability of a device for integration into work processes within socio-technical system-- these scales are subjective and roboticists may unknowingly overstate the maturity [13, 14]. Addressing this challenge impacts workforce training (e.g., should end-users be trained to anticipate problems? Should roboticists be trained to better predict readiness?), regulations (e.g., should regulations specify acceptable standards of trustworthiness?), and infrastructure (e.g., is there a cache of acceptable robots that can be safely used?).

While these challenges highlight the difficulties in accelerating the adoption of robots for infectious diseases, they suggest opportunities for the larger community. Both the end-user and robotics workforces require greater training, which, in turn, suggests a training process that synergistically brings together end-users and roboticists rather than trains separately. The challenges also suggest that

# The Role of Robotics in Infectious Disease Crises

appropriate regulations can incentivize the development of formal risk models that are acceptable to end-users and can inform agencies when to relax regulations in order to exploit when the benefit of robotics exceeds the risks. The challenges also expand the notion of infrastructure, both in terms of what end-users need to be able to cope with disaster and what roboticists need in order to develop robots that can be seamlessly translated to effective use.

## Regulatory and Reimbursement Barriers

The spectrum of uses of robots for the COVID-19 pandemic consisted of four general use cases: i) uses that were subject to regulations as to the minimum standards of safety and reliable operation (e.g., robots serving as medical devices for clinical applications, airworthiness of drones), ii) regulations prescribing whether a robotic device can be used for a particular application (e.g., telehealth, state and local regulations on the use of drones for surveillance), iii) applications that did not require regulatory consent (e.g., disinfection of businesses and schools, warehouse automation), and iv) applications that likely should have been subject to regulations (e.g., use of drones to identify and track infected citizens using unproven sensing technologies and without controls for privacy).

Currently most institutions and agencies have processes for waiving policies or regulations that would prevent the rapid introduction of robotics innovations. But the workshop noted concerns that they may not choose to do so. One reason may be the lack of quantified projection of risk of failures and unintended consequences including negative impacts on work processes. For example, waivers, especially those for clinical applications, rely on the primary stakeholder's judgment of the readiness of the technology for a particular application. For example, a hospital institutional review board can determine locally that the hospital can accept the risk of using a robot for non-medical purposes. A second reason why an institution may not adopt a technology is that it may not be reimbursable, even if it may be regulatorily permissible (full or with a waiver). Hospitals procedures for adopting new non-medical device technology often depend on whether insurers will cover the cost of acquisition or operation in addition to regulatory compliance. For example,, the regulations on the use of telehealth options for medical diagnosis and healthcare were quickly waived in Texas during the initial stages of the COVID-19 pandemic. But pervasive use occurred only after insurers agreed to accept those costs. The issue of reimbursable costs goes beyond healthcare; FEMA will not reimburse search and rescue teams or other responders for robotics, such as the use of drones. This second reason may be related to the first: if there existed quantitative methods for projecting risks and consequences, then robotics technologies might be more acceptable.

The principles of responsible innovation and professional ethics necessitate that waivers of regulations be driven by evidence, not emotion, and that new uses or new robots may require proactive regulations. One example is the use of thermal cameras on drones to detect and track infected citizens; there are no studies of the accuracy and indeed, the lack of success in using thermal cameras for the SARS and MERS outbreak strongly argued that this is not useful. As another example, rapidly produced clones of existing robots may not perform to the same levels as the original robot but end-users may assume that they are identical.

## Training Barriers

The introduction of robotics for novel uses, the increased use of existing robots, and the evaluation of the suitability of waiving regulations for responding to a pandemic indicate the need for end-user workforce training. This training takes two forms: *just-in-time training* and training incorporated into the normative vocational education and life-long certification processes. Of the two forms of training, just-in-time training will likely have the most impact but necessitates advances in human-robot interaction, as discussed in Section 3. Training should not be limited to end-users, as the COVID-19 pandemic

# The Role of Robotics in Infectious Disease Crises

suggests the advantage of training the robotics community to work with end-users and to understand demand analysis and regulatory constraints in order to effectively transfer advances in technology.

In general, end-users will require just-in-time training when robotic devices not in common use in their workplace are introduced. The time and circumstances of that training pose significant challenges. For example, even in ideal times, it would be unlikely to have nurses and staff take lengthy multi-day training on robots; they would not have much time during a pandemic. In addition, these end-users will likely already be working at capacity handling a surge and functioning under personal physiological and psychological stress from the event and long hours, which is not an ideal time for training. Indeed, with the notable exception of the Fukushima Daiichi nuclear accident, robots were operated by robot experts, not the true end user. In the case of Fukushima, TEPCO insisted on training its own personnel to use the robots which took a month; in hindsight, this is generally considered by all to have been a mistake. Note that the constraints on training an end-user implies that the user experience design of the robot, especially how easy it is for an end-user to quickly learn and get the robot operational, is crucial.

As the development of effective just-in-time training requires scientific advances in human-robot interaction, it may be more productive to focus on two other areas in the near term. One is training stakeholders with acquisition or regulatory roles in the disaster in advance so that they can conduct a demand analysis should a disaster occur. The other is on training roboticists on the work domain so that they can better assist with the demand analysis and better design user experience to reduce the need for lengthy training.

## Other Infrastructure Barriers

Infrastructure can connote either a) the state of robotic systems and their suitability, and availability, to be deployed and adapted to meet emerging demand pulls during a disaster or b) the physical or virtual infrastructure for conducting high fidelity research, testing and evaluation, and hands-on training. Both are important. However, prior studies, especially the NSF/CCC CRICIS report [4], has described the types of research infrastructure needed for university-user partnerships and high fidelity facilities to conduct formative experimentation and evaluation of progress and these will not be reviewed here.

The suitability and availability of robots to be deployed for a disaster are separate issues, with the suitability being the purview of the robotics community and the availability depending on economic and regulatory issues. As noted earlier, the suitability of robots for a disaster is largely driven by demand pull. In order to be used, robots must either already exist or be easily adaptable. For example, out of 203 instances of robots being used for 30 different applications for the COVID-19 pandemic, only 13 were for new missions. The rest were for missions that were well-established and robots already existed that were capable of performing the necessary functions.

There are three approaches to solving the availability problem: 1) creating and maintaining caches of robots; 2) exploiting or adding incentives for everyday adoption of robots so that they will be in use prior to the disaster; and 3) creating open source, vetted designs that can be quickly manufactured from low cost available robots or parts. The first approach of creating and maintaining caches of robots in anticipation of a disaster has not been successful historically. The robots tend to become quickly outdated, and no one maintains them or is trained on them. The second approach is interesting because many of the robots used for COVID-19 response were already in everyday use in clinical applications and public safety. Prior to the pandemic, many hospitals worldwide had disinfection and telehealth robots, which offered health benefits and economic savings. The UVC disinfection robots were motivated by the Ebola outbreak but are used routinely in hospitals to reduce hospital acquired infections. The third approach is also interesting because it was successfully taken by the Italian Institute of Robotics and Intelligent Machines (I-RIM). They created plans and software for telerobotics that could be assembled from an iRobot Roomba™ and parts that could be found at a hardware store. A challenge for open

# The Role of Robotics in Infectious Disease Crises

sourcing is a structure for maintaining and vetting designs and for ensuring no violations of intellectual property.

## 5. Recommended follow-on steps to develop a National strategic plan

This two-day workshop focused primarily on the potential of robotic systems to help meet clinical care and public health challenges such as those posed by the current COVID-19 pandemic. The discussion clearly indicated that these systems can play an increasingly important role in the current and future crises, as well as in non-crisis situations.

The effectiveness of any future response will depend significantly on development and implementation of a suitable strategic plan both to address key knowledge/technology barriers and to promote technology maturation to deployment. The strategic plan also needs to address significant infrastructure barriers that may impede rapid deployment of systems in future crises. Programs addressing these barriers must be married with programs that engage and mature end-user adoption of these technologies and ensure that there is continuous training and human resource development so that we are prepared for the next pandemic.

Although the development of such a strategy was beyond the scope of this limited workshop, it does provide some pointers for future development. Some immediate follow-on steps include:

- **Conduct a full NAE consensus study** on the potential for robotic systems to assist in healthcare emergencies and to develop a National Strategy for increasing national preparedness to use robotic systems and technology in future emergencies. This committee should include representatives from the engineering/robotics community, clinicians & critical care workers, public health experts, hospital epidemiologists, infectious diseases specialists, and emergency responders. It probably should operate under the auspices of, or in coordination with, the existing National Academies' Standing Committee on Emerging Infectious Diseases and 21st Century Health Threats, but with a specific focus on the role of robotics and related technologies.

- **Conduct scenario planning exercises** to "game out" future pandemic crises in order to identify areas where better preparation can facilitate more effective deployment of robotic systems to meet emergent needs. Such exercises could naturally be part of a full consensus study, but they could also be conducted independently. There should be multiple stakeholder-directed exercises reflecting the breadth of robotic applications for pandemics including: acute care, activities of daily living under quarantine, and medical supply chain; public safety; and industry and business.

- **Increase research addressing knowledge barriers affecting the capability and mobilization potential** of robotic systems, with **additional focus** on how these barriers might apply to robotic responses to infectious disease crises. Although of this research can be accommodated within the scope of existing programs, the use of clinical and nursing care as "use case" driving applications could encouraged.

# The Role of Robotics in Infectious Disease Crises- **Initiate targeted research programs addressing environment-specific knowledge barriers** to the deployment of robotic systems in infectious disease settings, as discussed in Section 3. Initiatives here could either be targeted toward specific barriers (e.g., materials used for construction of robots, disinfection of robot systems, or infectious agent detection) or could focus more broadly on complete systems (see below).

- **Develop incentives for translational research to develop novel robotic systems for healthcare and disaster applications.** Although research addressing fundamental knowledge barriers is *necessary*, it is not *sufficient* to close the gap between what is possible and what is practically deployable. Additional emphasis on complete systems for targeted applications can help close this gap while also providing sharper focus on relevant knowledge barriers. Training for roboticists to work in these domains with end-users as per 4.B. is important. Similarly, incentives for accelerating the transition from research prototypes to replicated and deployed systems for emerging crisis applications should be developed.

- **Develop training programs that expose practitioners to robotic technology.** There is a significant lack of awareness of how robots work and how they can be used in clinics and hospitals. We recommend the development of training modules and courses in robotics for medical schools and nursing schools, and training programs at the interface of engineering and medicine to promote this field.

- **Institute economic and policy incentives to accelerate adoption of adaptable robotic systems into everyday use by public agencies, particularly healthcare and the medical supply chain industry.** Policy incentives (particularly those impacting allowable reimbursable costs) are important, as are other economic incentives. A significantly larger installed base of robotic systems in these areas could be invaluable in responding to the next crisis, especially if these systems are designed in ways to promote ready adaptation to new circumstances.

- **Establish a consortium focused on robotic systems in communicable diseases** to facilitate connections between roboticists, government, industry, and citizen stakeholders and to provide a clearing house/repository for validated system designs and solutions that can be shared within the community. Such a consortium would provide a forum connecting the robotics community with the real needs, encourage groups to form among researchers, and prevent unproductive duplication of efforts. It would also enable researchers to overcome a lack of domain knowledge, especially for clinical care and public safety. And it would provide mechanisms to flatten the steep learning curve in understanding regulations.

## 6. Concluding Remarks

Pandemics can profoundly affect all aspects of our society, as the current COVID-19 outbreak has amply demonstrated. The goal of this workshop was to explore the roles that robots can play in addressing the many challenges presented by such crises. Although it was organized on relatively short notice and was conducted via videoconferencing over only two days, it nevertheless produced useful insights that could provide the basis for further study and action.

The discussion centered largely on the potential roles for robotic systems in areas directly related to health care applications. This is an emerging area and one that will be vitally important in dealing with future crises. The discussion identified a number of the knowledge and infrastructure barriers associated with healthcare applications. Robotic systems, generally, are becoming more "intelligent" and broadly capable, and many of these barriers are common across a very broad spectrum of application areas beyond health care. Although there is a great deal of synergy between robotic

# The Role of Robotics in Infectious Disease Crises

systems for health care and those for more generalized use, there are also some specific barriers that also need to be addressed.

The discussion also identified several important follow-on steps that are summarized in Section 5 of this report. Perhaps the most important recommendation is that a much more thorough and comprehensive study to develop a coherent National Strategy be undertaken for increasing national preparedness to use robotic systems and technology in future emergencies. However, there are a number of actions that can be taken concurrently that will also increase our preparedness while also promoting the more general development of highly capable robotic systems to fulfill many important societal needs.

Finally, the organizers of this workshop express our appreciation to the NAE and CCC for their sponsorship of this workshop and our hope that this report may prove useful.

# The Role of Robotics in Infectious Disease Crises

# Appendices

## A. Workshop Participants

| Name | Last | Affiliation |
|---|---|---|
| Gopika | Ajaykumar | Johns Hopkins University |
| Tamara | Barboza | Boston University |
| Antonio | Bicchi | Istituto Italiano di Tecnologia & Università di Pisa |
| George | Demiris | University of Pennsylvania |
| Bill | Brody | Retired |
| Andy | Ding | Johns Hopkins University School of Medicine |
| Irina | Dolinskaya | NSF |
| Khari | Douglas | CRA/CCC |
| Zeke | Emanuel | University of Pennsylvania |
| Elazer | Edelman | MIT |
| Jason | Farley | Johns Hopkins School of Nursing |
| Vignesh Babu Manjunath | Gandudi | Texas A&M University |
| Ken | Goldberg | UC Berkeley |
| Jesse | Goodman | Georgetown University |
| Aryaman | Gupta | Johns Hopkins |
| Greg | Hager | Johns Hopkins University |
| Peter | Harsha | Computing Research Association |
| Kris | Hauser | University of Illinois at Urbana-Champaign |
| Ayanna | Howard | Georgia Institute of Technology |
| Louise | Howe | National Science Foundation |
| Karen | Jacobs | Boston University |
| Michelle | Johnson | University of Pennsylvania |
| Lydia | Kavraki | Rice University |

# The Role of Robotics in Infectious Disease Crises

| Axel | Krieger | Johns Hopkins University |
|---|---|---|
| Vijay | Kumar | University of Pennsylvania |
| James | Lawler | U. Nebraska Medical Center |
| Brian | Litt | University of Pennsylvania |
| Daniel | Lopresti | Lehigh University |
| Alexander | Lu | JHU |
| Guru | Madhavan | National Academy of Engineering |
| Margaret | Martonosi | NSF |
| Robin | Murphy | Texas A&M University |
| Wendy | Nilsen | National Science Foundation |
| Gerald | Parker | Texas A&M University |
| Lynne | Parker | OSTP |
| Erion | Plaku | National Science Foundation |
| Sohi | Rastegar | National Science Foundation |
| Daniela | Rus | CSAIL |
| Ann | Schwartz Drobnis | CRA/CCC |
| Nirav | Shah | Stanford University |
| Erica | Shenoy | Massachusetts General Hospital and Harvard Medical School |
| Russell | Taylor | Johns Hopkins University |
| Madeleine | Waldram | The Johns Hopkins University |
| Helen | Wright | CRA/CCC |

## B. Workshop Agenda

**July 9, 2020 (Thursday)**

| 11:00 AM | **Welcome / Introductions / Opening** |
|---|---|

# The Role of Robotics in Infectious Disease Crises

| | |
|---|---|
| 11:30 AM | **Keynote**<br><br>*Ezekiel Jonathan "Zeke" Emanuel is an American oncologist, bioethicist and senior fellow at the Center for American Progress. He is the current Vice Provost for Global Initiatives at the University of Pennsylvania and chair of the Department of Medical Ethics and Health Policy.* |
| 12:00 PM | **Stage Setting Talks- A**<br><br>What are the needs and roles for robotics in infectious disease crises such as the current COVID-19 pandemic?<br><br>*James Lawler (University of Nebraska Medical Center)*<br><br>*Antonio Bicchi (Istituto Italiano di Tecnologia and University of Pisa)* |
| 12:30 PM | **Breakout A** |
| 01:15 PM | **BREAK** |
| 01:45 PM | **Report Back A** |
| 02:30 PM | **Stage Setting Talks- B**<br><br>What are the technology and technical readiness challenges in meeting those needs?<br><br>*George Demiris (University of Pennsylvania)*<br><br>*Ken Goldberg (University of California, Berkeley)* |
| 03:00 PM | **Breakout B** |
| 03:45 PM | **BREAK** |
| 04:15 PM | **Report Back B** |

# The Role of Robotics in Infectious Disease Crises

| | |
|---|---|
| 04:45 PM | **Wrap Up Day 1** |

**July 10, 2020 (Friday)**

| | |
|---|---|
| 11:00 AM | **Day 2 Welcome/Recap** |
| 11:05 AM | **Stage Setting Talks- C**<br>What is the role of research, industry, and government in accelerating readiness for the next crisis?<br>Lynne Parker (OSTP)<br>Nirav Shah (Stanford)<br>Bill Brody (Former President of Salk Institute) |
| 11:35 AM | **Breakout C** |
| 12:35 PM | **BREAK** |
| 01:00 PM | **Report Back C** |
| 01:45 PM | **Full Group Conversation** |
| 02:15 PM | **WRITING** |

## C. Typical COVID-19 Projects

**These are some of the many projects currently underway with robots working in the current COVID-19 pandemic. They are listed here as an example of the type of work currently being carried out, not as an exhaustive list.**

1. **Corus Robotics** rolling UV bots disinfects offices. See **Boston Globe 9/7**: https://www.bostonglobe.com/2020/09/07/business/office-bots-rolling-uv-machine-is-disinfecting-workplaces
2. MIT and Ava Robotics on a UVC disinfecting robot at the Greater Boston Food Bank https://news.mit.edu/2020/csail-robot-disinfects-greater-boston-food-bank-covid-19-0629
3. Johns Hopkins and University of Maryland have developed a simple telerobotic system for operating ventilators remotely from outside an infectious disease ICU. **JHU Hub article and video:**

# The Role of Robotics in Infectious Disease Crises

https://hub.jhu.edu/2020/08/13/remote-control-ventilators-covid-19-pandemic/ and **NBC Nightly News segment:** https://www.nbcnews.com/nightly-news/video/robots-now-on-the-front-lines-in-battle-against-covid-19-89608773646

4. Mobile robots used in hospital and grocery store logistics. **NBC Nightly News segment:** https://www.nbcnews.com/nightly-news/video/robots-now-on-the-front-lines-in-battle-against-covid-19-89608773646 (also includes segment on Johns Hopkins ICU robot)

5. More *Covid related* innovations. See **IEEE Spectrum 9/4**: https://spectrum.ieee.org/robotics & https://spectrum.ieee.org/static/covid19-ieee-resources

6. *Robot trucks* moving into logistics and freight. See **WS Journal 9/1**: https://www.wsj.com/articles/robot-trucks-are-seeking-inroads-into-freight-business-11598954400?mod=searchresults&page=1&pos=9

7. Amazon secures approval for *drone delivery*. See **WS Journal 8/31.** https://www.wsj.com/articles/amazon-gets-u-s-approval-for-drone-fleet-a-package-delivery-milestone-11598913514?mod=searchresults&page=1&pos=2

8. Farm worker shortage in Covid boosts *robots in farming*. See **Financial Times 8/30**. https://www.ft.com/content/0b394693-137b-40a4-992b-0b742202e4e1

9. **Boston Dynamics' "Spot"** robot featured in the Economist on feature on robots in factories. See **Economist 8/25:** https://www.economist.com/science-and-technology/2020/08/25/robots-that-can-walk-are-now-striding-to-market

10. **Root AI** raised $7.2 million in seed funding for its crop-picking robot. See **The Robot Report 8/14 & Boston Business Journal 8/14**: https://www.therobotreport.com/root-ai-raises-seed-funding-for-virgo-robot-designed-to-harvest-multiple-crops & https://www.bizjournals.com/boston/news/2020/08/14/form-d-friday-nanotech-robotics-energy-companies.html

11. **Brainavi** has developed a robotic application to perform nasal swabs for COVID-19 sample collection: https://brainnavi.com/product/nasalswabrobot/

12. A **volunteer organization** working in conjunction with the IEEE Robotics and Automation Special Interest Group on Humanitarian Technology (RAS-SIGHT) has set up a web site including interviews with experts and a survey of robotic systems deployed or being developed for the current COVID-19 crisis: http://roboticsforinfectiousdiseases.org; http://roboticsforinfectiousdiseases.org/how-robots-are-being-used.html.

### D. Selected Additional Articles on Robotic Systems for COVID-19

There is a large and growing literature on robotic systems developed for or applied to the current COVID-19 pandemic. Here are a few typical examples, in addition to the work cited in the references to the main report.

1. A. A. Malik, T. Masood, and R. Kousar, "Repurposing factories with robotics in the face of COVID-19", *Science Robotics*, vol. 5-, p. eabc2782, 2020. June 17 June 17. https://robotics.sciencemag.org/content/robotics/5/43/eabc2782.full.pdf   DOI: 10.1126/scirobotics.abc2782

2. A. Malik, T. Masood, and R. Kousar, "Reconfiguring and ramping-up ventilator production in the face of COVID-19: Can robots help?", arXiv preprint arXiv:2004.07360, 2020.

# The Role of Robotics in Infectious Disease Crises

# The Role of Robotics in Infectious Disease Crises